\documentclass[unnumsec,webpdf,contemporary,large]{oup-authoring-template}

\usepackage[T1]{fontenc}
\usepackage{amsmath}
\usepackage[utf8]{inputenc}
\usepackage{indentfirst}
\usepackage{hyperref}
\usepackage{algorithm}
\usepackage{algpseudocode}
\usepackage{graphicx}
\usepackage{multirow}
\usepackage{booktabs}  
\usepackage{colortbl}  
\usepackage{caption}   
\usepackage{booktabs}

\graphicspath{{Fig/}}

\theoremstyle{thmstyleone}%

\theoremstyle{thmstyletwo}%
\theoremstyle{thmstylethree}%

\hypersetup{
  bookmarksdepth=section
}
\begin{document}
\journaltitle{Briefings in Bioinformatics}
\DOI{DOI HERE}
\copyrightyear{2024}
\pubyear{2019}
\access{Advance Access Publication Date: Day Month Year}
\appnotes{Paper}

\firstpage{1}


\title[CSGDN]{CSGDN: Contrastive Signed Graph Diffusion Network for Predicting Crop Gene-phenotype Associations}

\author[]{Yiru Pan \ORCID{0009-0008-6773-2464}$^{\ddagger}$}
\author[]{Xingyu Ji \ORCID{0009-0002-5714-4489}$^{\ddagger}$}
\author[]{Jiaqi You \ORCID{0000-0001-6569-1955}$^{\ddagger}$}
\author[]{Lu Li \ORCID{0009-0000-6800-8373}}
\author[]{Zhenping Liu \ORCID{0000-0002-3046-8005}}
\author[]{Xianlong Zhang \ORCID{0000-0002-7703-524X}}
\author[$\ast$]{Zeyu Zhang \ORCID{0000-0002-2376-6151}}
\author[$\ast$]{Maojun Wang \ORCID{0000-0002-4791-3742}}

\authormark{Yiru Pan et al.}

\address[]{\orgdiv{National Key Laboratory of Crop Genetic Improvement}, \orgname{Hubei Hongshan Laboratory, Huazhong Agricultural University}, \postcode{430070}, \state{Hubei}, \country{China}}

\corresp[$\ast$]{Corresponding author. \href{email: mjwang@mail.hzau.edu.cn}{email: mjwang@mail.hzau.edu.cn}, \href{email: zhangzeyu@mail.hzau.edu.cn}{zhangzeyu@mail.hzau.edu.cn} \\
$^{\ddagger}$ Yiru Pan, Xingyu Ji and Jiaqi You contributed equally to this work.}

\received{Date}{0}{Year}
\revised{Date}{0}{Year}
\accepted{Date}{0}{Year}



\abstract{Positive and negative association prediction between gene and phenotype helps to illustrate the underlying mechanism of complex traits in organisms. The transcription and regulation activity of specific genes will be adjusted accordingly in different cell types, developmental stages, and physiological states. There are the following two problems in obtaining the positive/negative associations between gene and trait: 1) High-throughput DNA/RNA sequencing and phenotyping are expensive and time-consuming due to the need to process large sample sizes; 2) experiments introduce both random and systematic errors, and, meanwhile, calculations or predictions using software or models may produce noise. To address these two issues, we propose a \underline{\textbf{C}}ontrastive \underline{\textbf{S}}igned \underline{\textbf{G}}raph \underline{\textbf{D}}iffusion \underline{\textbf{N}}etwork, \textbf{CSGDN}, to learn robust node representations with fewer training samples to achieve higher link prediction accuracy. CSGDN employs a signed graph diffusion method to uncover the underlying regulatory associations between genes and phenotypes.  Then, stochastic perturbation strategies are used to create two views for both original and diffusive graphs. Lastly, a multi-view contrastive learning paradigm loss is designed to unify the node presentations learned from the two views to resist interference and reduce noise. We conduct experiments to validate the performance of CSGDN on three crop datasets: \textit{Gossypium hirsutum}, \textit{\textit{Brassica napus}}, and \textit{Triticum turgidum}. The results demonstrate that the proposed model outperforms state-of-the-art methods by up to 9.28\% AUC for link sign prediction in \textit{G.\ hirsutum} dataset. The source code of our model is available at \url{https://github.com/Erican-Ji/CSGDN}.}
\keywords{gene-phenotype associations, graph neural networks, signed bipartite networks, signed graph neural networks}

\maketitle

\begin{figure*}[ht]
    \centering
    \includegraphics[width=\textwidth]{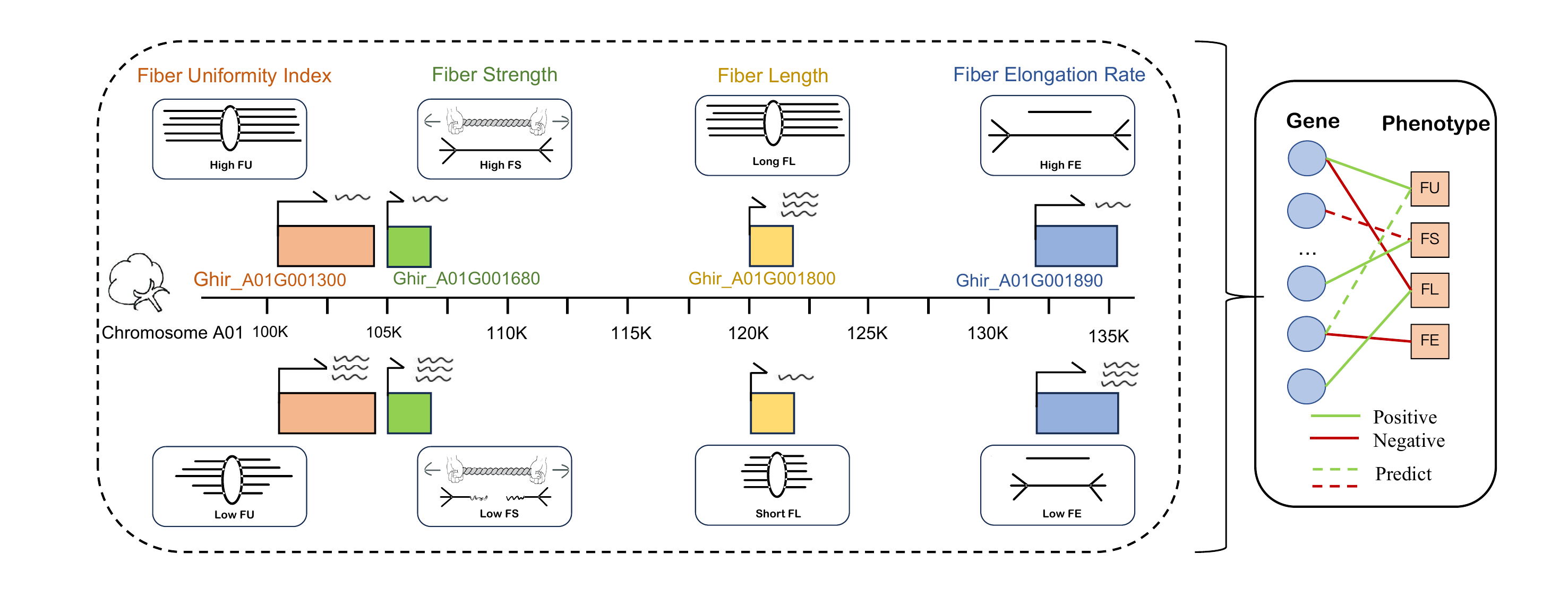}
    \caption{CSGDN abstracts the associations between genes and phenotypes into a signed bipartite graph. Our task is to predict the gene-phenotype associations by constructing a neural network framework for the bipartite graph.}
    \label{fig:introduction}
\end{figure*}

\section{INTRODUCTION}

Positive/negative association prediction between gene and phenotype has long been known to be important in the field of biology, with broad applications in breeding crops \cite{li2017natural}. By constructing the association between genomic variations (such as single nucleotide polymorphisms, SNPs) and phenotypes within a large population, genome-wide association study (GWAS) uncovers the genetic basis underlying the formation of the phenotypes \cite{visscher201710, pasaniuc2017dissecting, tam2019benefits, qi2024genetic}. In many crops such as rice, maize, wheat, cotton and rapeseed, GWAS has been widely applied to identify quantitative phenotype loci (QTLs) related to yield \cite{liu2023high, lin2024systematic, zhang2023ggamma, miller2019variation, ma2018resequencing}, quality \cite{ma2018resequencing, you2023regulatory, zhao2022genomic, wang2022genomic}, resistance \cite{zhang2023ggamma, zhao2022genomic, tian2023allelic, wang2024tabhlh27, li2020phenomics, ma2021combination}, and other agronomic phenotypes \cite{miller2019variation, zhang2024integrating, tang2021genome}, and to further help identify candidate genes. In sorghum, an alkaline tolerance related gene named \textit{Alkali Tolerance 1 (AT1)} was located through GWAS analysis of 352 representative accessions, which encodes a heterotrimeric G protein $\gamma$ subunit (G$\gamma$) [10]. The nonfunctional mutant can improve the yield in sorghum, rice and maize in alkaline soils. In rapeseed, Zhang et al. (2024) located a previously unknown gene on chromosome A02, named \textit{BnaA02.SE} \cite{zhang2023ggamma}. Overexpression of this gene contributes to longer and larger siliquae, indicating that the gene is positively associated with rapeseed yield.

Compared with validating candidate genes through molecular biology approaches, the strategies integrated transcriptome such as GWAS-eQTL colocalization methods (eg. coloc, eCaviar) \cite{giambartolomei2014bayesian, hormozdiari2016colocalization} and mendelian randomization methods (eg. SMR/HEIDI) \cite{zhu2016integration, commonmind2017erratum} assist with evaluating gene-phenotype associations at an earlier stage, which narrow down the number of candidate genes and greatly reduce the time and cost required for creating mutant materials. Transcriptome-wide association study (TWAS) builds expression prediction models based on BLUP (Best Linear Unbiased Prediction) \cite{robinson1991blup} and BSLMM (Bayesian Sparse Linear Mixed Model) \cite{zhou2013polygenic}, enabling the prediction of large-scale expression levels from small-scale transcriptomes, which further addresses challenges such as difficultly or costly sampling for RNA sequencing \cite{zhu2016integration, wainberg2019opportunities}. However, the number of genes that can be accurately imputed is still limited by the training cohort size and the quality of the training data \cite{gusev2016integrative}; and TWAS strategy still relies on the framework of association analysis, lacking accurate predictive ability for genes with low variant density.

Additionally, traditional methods can introduce various types of errors and noise. For instance, biases such as systematic bias, coverage bias, and batch effects often affect Next-Generation Sequencing (NGS) data \cite{slatko2018overview}. These biases are introduced by sequencing platform, genome content and experimental variability \cite{taub2010overcoming}. Experimental design and the selection of sample replication depend on the specificity of the species, which to some extent affects the bias of quantitative results \cite{conesa2016survey}. The accurate TWAS results \cite{gusev2016integrative} also depend on the high quality of training data. Therefore, it is crucial to develop models to predict positive/negative gene-phenotype associations, minimizing the biases and errors produced in traditional experimental methods.

As mentioned, there are two major issues in predicting the positive/negative regulation associations between gene and phenotype:

(1) \textbf{The high cost associated with large sample sizes.}  
Traditional methods like GWAS and TWAS require large sample sizes to detect associations, which increases costs for sequencing and data analysis. Larger samples also demand more time and resources, making the process expensive and slow.

(2) \textbf{Data noise from experiments and methods.} Experiments often introduce biases such as systematic errors and batch effects. Noise makes it difficult to accurately detect the associations between gene and phenotype.

In response to these two issues, we propose a \underline{\textbf{C}}ontrastive \underline{\textbf{S}}igned \underline{\textbf{G}}raph \underline{\textbf{D}}iffusion \underline{\textbf{N}}etwork, \textbf{CSGDN} to solve these issues of predicting the positive/negative regulation associations. In this approach, gene-phenotype associations are modeled as a signed bipartite graph, where gene and phenotype nodes are connected by either positive or negative edges, indicating positive/negative-regulation of genes in phenotypes. \textbf{For obstacle 1}, we applies a signed graph diffusion theory to uncover hidden associations between genes and phenotypes. The diffusion method helps address the challenge of large sample size requirements by capturing complex associations through a smaller dataset, reducing the overall cost. \textbf{For obstacle 2}, we employs stochastic perturbation techniques to generate two views of both the original and diffused graphs. A multi-view contrastive learning loss unifies the node representations from both views, helping to reduce interference and noise. In summary, our model effectively addresses the cost and noise challenges as mentioned. CSGDN can predict gene-phenotype associations across crop species using only small-sample associations, and shows strong robustness against interference from experimental noise.

To evaluate the effectiveness of CSGDN, we conduct extensive experiments on three crop datasets: \textit{Gossypium hirsutum}, \textit{Brassica napus}, and \textit{Triticum turgidum}. Our results demonstrate that CSGDN consistently outperforms baseline models from both unsigned GNNs and signed GNNs. For unsigned GNNs (GCN, GAT, GRACE), CSGDN achieves improvements in AUC of up to 6.66\%, 7.13\%, and 7.26\%, respectively. For signed GNNs (SGCN, SGCL, SGNNMD), CSGDN outperforms the baselines with AUC gains of 5.96\%, 5.29\%, and 9.29\%, respectively. These results underscore CSGDN’s superior ability to enhance link sign prediction tasks across diverse crop datasets. We randomly sample 80\% of the datasets to evaluate CSGDN’s performance in addressing the challenge of small sample sizes. The results show that our model outperforms the baselines on AUC, Binary-F1, Micro-F1, and Macro-F1, with improvements of up to approximately 10\% across all metrics. Specifically, CSGDN enhances Binary-F1 by up to 9.51\% with 10\% perturbed edges and improves Micro-F1 by up to 12.64\% with 20\% perturbed edges compared to SGNNMD. These experiments clearly highlight the effectiveness of our model CSGDN.

Overall, our contributions are summarized as follows:
\begin{itemize}
    \item We propose a model for gene-phenotype association prediction, which outperforms baselines in terms of link sign prediction performance.
    \item By combining the diffusion method and the contrastive learning framework, our model effectively addresses the challenges of high cost and noise in positive/negative regulation association prediction.
\end{itemize}

\section{RELATED WORK}

\textbf{Association Prediction}. In recent years, significant progress has been made in the field of association prediction. SGNNMD proposed by Zhang et al. \cite{zhang2022sgnnmd}, extracts subgraphs around miRNA-disease pairs from the signed bipartite networks and utilizes biological features to accurately predict deregulation types of miRNA-disease associations. The framework HGATLDA proposed by Zhao et al. \cite{zhao2022heterogeneous}, effectively integrates meta-path-based heterogeneous graphs and attention mechanisms to improve the prediction of lncRNA–disease associations, addressing limitations in feature fusion, complex associations, and data imbalance. MTRD proposed by Zhang et al. \cite{zhang2022learning} integrates multi-scale topology embeddings and node attributes with advanced learning mechanisms, achieving superior performance in predicting drug-disease associations and effectively identifying potential candidate diseases for specific drugs. BGF-CMAP proposed by Guo et al. \cite{guo2024biolinguistic} integrates advanced techniques like Word2vec and graph embedding algorithms to extract both sequence and interaction features, significantly enhancing the prediction of complex circRNA–miRNA associations. MDGF-MCEC proposed by Wu et al. \cite {wu2022mdgf}, utilizes a multi-view dual attention GCN and cooperative ensemble learning to predict circRNA-disease associations, achieving enhanced feature representation and classification through multi-similarity relation graphs and attention-based feature adjustment. Although neural networks have shown great potential in various association prediction fields, there is still insufficient progress in developing direct prediction methods for gene-phenotype associations, particularly due to the high costs associated with large sample sizes and the noise introduced throughout the experimental processes.

\textbf{Signed Graph Neural Networks}. Recently, neural approaches have gained traction in signed graph representation, we refer to such methods collectively as Signed Graph Neural Networks (SGNNs) \cite{derr2018signed,zhang2023contrastive,li2024se,he2024mitigating,zhang2024dropedge}. With innovations like SiNE \cite{wang2017signed} pioneering deep learning use by leveraging triangle structures with mixed edge polarities. SiNE optimizes an objective grounded in balance theory for embedding generation. The advent of SGCN \cite{derr2018signed} extended GCN's \cite{kipf2016semi} scope to signed graphs, incorporating balance theory for multi-hop relationship discernment. Models like SiGAT \cite{huang2019signed}, SNEA \cite{li2020learning}, SDGNN \cite{huang2021sdgnn}, and SGCL \cite{shu2021sgcl}, rooted in attention mechanisms \cite{vaswani2017attention}, further enrich this landscape. 

Beyond the aforementioned approaches, some SGNNs claim to possess robustness against noise. SGCL \cite{shu2021sgcl} first adopts contrastive learning paradigm in signed graph analysis, which can learn invariant representation of nodes under minor random perturbation and help to enhance the robustness of the model. RSGNN \cite{zhang2023rsgnn} is another trial to enhance the robustness of SGNNs. It enables the model to learn a cleaner structure, less influenced by noise. However, considering the substantial computational overhead involved in learning a new adjacency matrix, this approach is only feasible for smaller node sets. Given the vast number of gene types, this method is impractical for gene-phenotype datasets. 

\section{MATERIALS}

\subsection{DataSets}

We use datasets from three species including \textit{G.\ hirsutum} \cite{you2023regulatory}, \textit{\textit{B.\ napus}} \cite{tang2021genome} and \textit{T.\ turgidum} \cite{yang2024combined} to input our model. For each species data, we perform TWAS process method to obtain associations between gene and phenotype. Initially, phenotype data is standardized using qqnorm function in R and Principal Component Analysis (PCA) is conducted for population structure data using TASSEL software \cite{bradbury2007tassel}. FaST-LMM software \cite{lippert2011fast} is employed to perform GWAS, considering phenotype data and population structure. We use Fusion software for performing TWAS \cite{gusev2016integrative} to obtain TWAS $z score$ of associations between partial genes and phenotypes. The positive or negative sign of TWAS $z score$ indicates positive/negative regulation, respectively. Note that TWAS process can only calculate partial associations between genes and phenotypes. We separate the genes that can be associated with TWAS from those that cannot be calculated by TWAS within the species, and input them separately into the model.

\textbf{\textit{G.\ hirsutum}} \cite{you2023regulatory}. The phenotypes data include four phenotypes at the 4 DPA (days post anthesis) stage: Fiber length (FL), Fiber strength (FS), Fiber elongation rate (FE), Fiber Uniformity (FU). The reference genome is from Wang et al. \cite{wang2019reference}. The result of TWAS process contains 523 associations between genes and FL phenotype, 1129 associations between genes and phenotype FS, 1521 associations between genes and phenotype FE, 509 associations between genes and phenotype FU.

\textbf{\textit{B.\ napus}} \cite{tang2021genome}. \textit{B.\ napus} dataset includes 20 DPA and 40 DPA stages. phenotype type is only seed oil content (SOC). The reference genome (\textit{B. napus} v4.1) can be downloaded from Genoscope (http://www.genoscope.cns.fr/brassicanapus/). The number of associations between genes in \textit{B.\ napus} and SOC are 605 at the 20 DPA stage and 148 at the 40 DPA stage, separately. We regard the different stages as distinct phenotypes.

\textbf{\textit{T.\ turgidum}} \cite{yang2024combined}. \textit{T.\ turgidum} dataset includes four phenotypes: Biomass (BM), Root length (RL), Root area (RA), and Root volume (RV). We use WEW v2.1 as the reference genome \cite{muslu2021comparative}. The result of the TWAS process includes 36 associations with these four phenotypes, with BM having 1 association, RA having 2, RL having 17, and RV having 15.

\subsection{Gene Sequences Similarity}

As the complement to our model, we calculate the similarity between gene sequences as features. The software BLAST+/2.9.0 \cite{camacho2009blast+} is used for aligning sequences between gene pairs. BLAST can calculate the similarity scores between gene pairs. 
We use BLAST+ 2.9.0 \cite{camacho2009blast+} to obtain a similarity matrix $S \in \mathbb{R}^{n \times n}$, where $n$ = $|\mathcal{V}|$. 
We use the similarity matrix $S$ as the initial feature input for genes. To be specific, 
the input feature $h_{i}^{(0)}$ of gene $v_i$ is the $i$-th row of the similarity matrix.

\section{Preliminary}

\begin{table}[]
\centering
\caption{Main notations used throughout this paper.}
\label{tab:notations}
\resizebox{0.47\textwidth}{!}{ 
\begin{tabular}{p{1.3cm} p{6cm}}
\toprule
    \textbf{Notations} & \textbf{Descriptions} \\
\midrule
    $\mathcal{G}$ & A signed graph \\
    $\mathcal{V}$ & A set of gene nodes \\
    $\mathcal{U}$ & A set of phenotype nodes \\
    $\mathcal{E}$ & A set of edges, where $\mathcal{E}^+$ denotes positive edges and $\mathcal{E}^-$ denotes negative edges \\
    $\mathcal{E}_{\text{TWAS}}$ & A subset of edges calculated through TWAS processes, defined as $\mathcal{E}_{\text{TWAS}} \subseteq \mathcal{E}$ such that $\forall (v, u) \in \mathcal{E}_{\text{TWAS}}$, where $v \in \mathcal{V}$ and $u \in \mathcal{U}$ \\
    $\mathcal{V}_{\text{TWAS}}$ & The set of gene nodes associated with TWAS, defined as $\mathcal{V}_{\text{TWAS}} = \{ v \in \mathcal{V} \,|\, \exists u \in \mathcal{U}, \ (v, u) \in \mathcal{E}_{\text{TWAS}} \}$ \\
    $\mathcal{S}$ & A similarity matrix of genes \\
    $\mathbf{A}$ & The adjacency matrix of the graph $\mathcal{G}$ \\
    $\mathbf{D}$ & The outdegree diagonal matrix of the graph $\mathcal{G}$ \\
    $\mathbf{G_k}$ & The graphs obtained by the graph augmentation method \\
    $h_{i,k}^{(l)}$ & The representation of the $i$-th node for the $l$-th layer in the $k$-th augmented graph \\
    $z_{i,k}^+$ & The final representation of the $i$-th node in the positive graph of the $k$-th augmented graph \\
    $z_{i,k}^-$ & The final representation of the $i$-th node in the negative graph of the $k$-th augmented graph \\
\bottomrule
\end{tabular}
}
\end{table}
We construct a signed bipartite graph  $\mathcal{G} = \{\mathcal{V},\mathcal{U}, \mathcal{\mathcal{E}}^+, \mathcal{E}^-\}$, where a matrix set of gene $\mathcal{V} = \{v_1, v_2, \ldots, v_{|\mathcal{V}|}\}$ and a set of phenotypes $\mathcal{U} = \{u_1, u_2, \ldots, u_{|\mathcal{U}|}\}$ are given. The edge set $\mathcal{E}$ represents deregulation associations, where $\mathcal{E} = \mathcal{E}^+ \cup \mathcal{E}^-$ and $\mathcal{E}^+ \cap \mathcal{E}^- = \emptyset$. Each edge belongs to one regulation type: a positive edge  $(v, u) \in \mathcal{E}^+$ indicates that the specific gene up-regulates the phenotype, while a negative edge $(v, u) \in \mathcal{E}^-$ represents down-regulation. 

We define a subset of the edge set $\mathcal{E}$ as $\mathcal{E}_{\text{TWAS}}$, which consists of edges that can be calculated through TWAS processes. Consequently, the node set $\mathcal{V}_{\text{TWAS}}$ is defined as the set of genes that correspond to the edges in $\mathcal{E}_{\text{TWAS}}$: $\mathcal{E}_{\text{TWAS}} \subseteq \mathcal{E} \quad \text{such that} \quad \forall (v, u) \in \mathcal{E}_{\text{TWAS}}$, where $v \in \mathcal{V}, u \in \mathcal{U}$. The gene set associated with TWAS can then be expressed as: $\mathcal{V}_{\text{TWAS}} = \{ v \in \mathcal{V} \,|\, \exists u \in \mathcal{U}, \ (v, u) \in \mathcal{E}_{\text{TWAS}} \}$

Given $\mathcal{G} = \{\mathcal{U}, \mathcal{V}, \mathcal{\mathcal{E}}^+, \mathcal{E}^-\}$, the goal is to learn a function $f$ to map nodes $v_i \in \mathcal{V}$ and $u_j \in \mathcal{U}$ to low-dimensional embeddings $z_{v_i} \in \mathbb{R}^d$ and $z_{u_j} \in \mathbb{R}^d$, where $d$ is the dimension of node embeddings. These embeddings should be useful for downstream tasks such as link sign prediction. This setup forms a crucial framework for studying the associations between genes and phenotypes, allowing for the prediction of up-regulation or down-regulation patterns between genes and phenotypes.

\section{PROPOSED METHOD}

\begin{figure*}[ht]
    \centering
    \includegraphics[width=\textwidth]{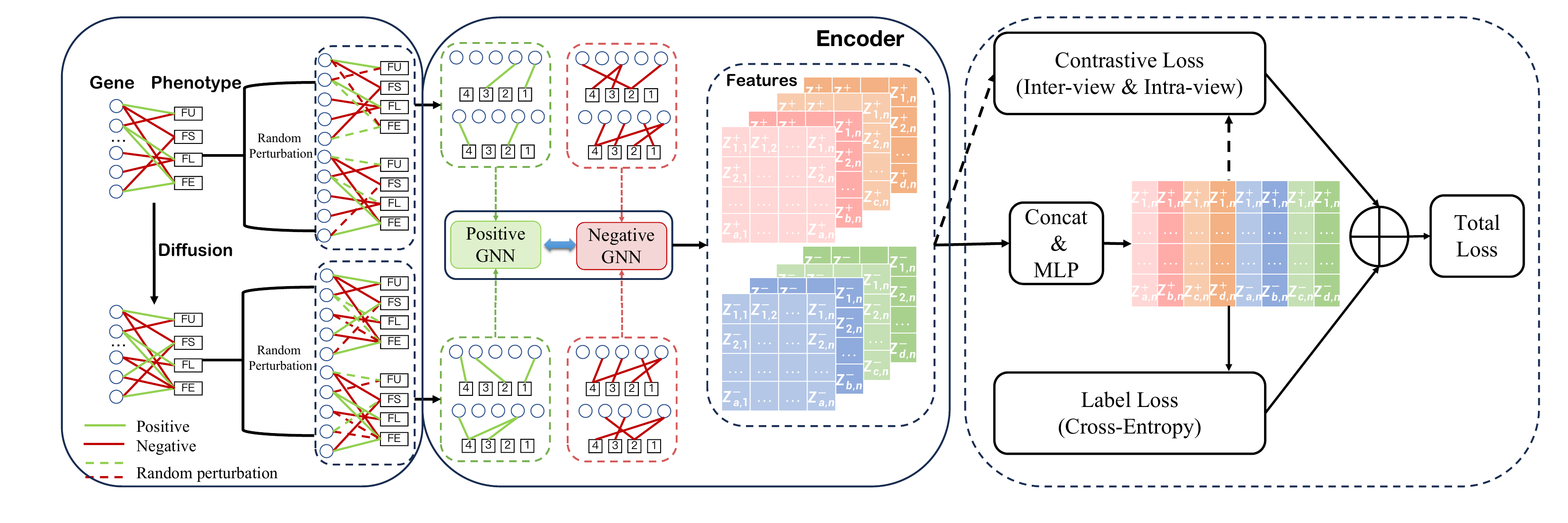}
    \caption{The overall architecture of CSGDN.}
    \label{fig:workflow}
\end{figure*}

In this section, we introduce the \underline{\textbf{C}}ontrastive \underline{\textbf{S}}igned \underline{\textbf{G}}raph \underline{\textbf{D}}iffusion \underline{\textbf{N}}etwork (\textbf{CSGDN}) model to tackle the aforementioned challenges, including enhancing structural information for small samples to reduce the expense in TWAS analysis and reducing noise from the whole process. As mentioned in MATERIALS section, we divide genes in each species into two parts: one part is associated with phenotypes through the TWAS process, and the other part lacks such associations. 

The main framework of the proposed CSGDN model is to encode genes associated with TWAS and phenotypes, which is illustrated in Fig.2. The main framework includes four key components: (1) graph diffusion; (2) graph augmentation; (3) the encoder; and (4) contrastive learning.
To be specific, we first use a graph diffusion method to uncover the potential relationships between genes and phenotypes, resulting in a diffusion graph. Then, we randomly remove edges from the original graph and the diffusion graph to obtain augmented graphs. After encoding these augmented graphs with the encoder, we obtain node representations to calculate the contrastive loss. 

In addition, we train a MLP to learn the encoding capability of the model's main framework for TWAS genes, in order to encode genes that lack TWAS associations.

\subsection{Signed Graph Diffusion}
The collection of large-scale crop sample data requires a high cost. Therefore, developing methods that can maintain high-precision prediction even with a low number of training samples is very necessary. In the field of biology, the effects of genes on phenotypes involves both positive and negative regulation, meaning that genes can either promote or inhibit phenotypes through complex mechanisms. This forms a bipartite signed graph, which consists of positive and negative edge types and two disjoint sets of nodes. In situations where crop sample data is scarce, the resulting signed graph structure is relatively sparse.

Traditional ranking models, such as PageRank \cite{page1999pagerank}, can mine the potential relationships between nodes to increase the number of edges but are only suitable for graphs with a single type of positive edges. Wu et al. \cite{wu2016troll} proposed Troll-Trust model (TR-TR) which is a variant of PageRank, without explanation of complex
relationships between negative and positive edges. To better handle the signed graph with two edge types, we use a diffusion operation based on the Signed Random Walk with Restart (SRWR) algorithm proposed by previous researchers to uncover the potential associations between genes and phenotypes, obtaining a diffusion graph.  
The diffusion graph provides richer structural information, which can effectively alleviate the problem of sparsity in graph data, while also providing a new graph structure for subsequent contrastive learning operations.

To be specific, the diffusion method is based on the balance theory. In balance theory, there are four types of relationships between nodes: friend's friend, friend's enemy, enemy's friend, and enemy 's enemy. On this basis, we introduce a signed random surfer for bipartite signed graph. The surfer randomly surfs between nodes and traverses their relationships. When the surfer encounters a positive edge, it maintains a positive sign +; when it encounters a negative edge, it flips its sign to negative -. Initially, the surfer carries a positive sign + at node \( u \). If the surfer is at node \( v \) at this time and the restart probability is \( c \), the probability of the surfer randomly surfing to neighboring nodes is \( 1 - c \).

\( \mathbf{r}_u^+ \) is the probability that a positive sign surfer is at node \( u \) from the seed node \( s \), and \( \mathbf{r}_u^- \) is the probability that a negative sign surfer is at node \( u \) from the seed node \( s \). We can formulate \( \mathbf{r}_u^+ \) and \( \mathbf{r}_u^- \) by Equation 1 below where \(\overleftarrow{\mathbf{N}}_{u}^{+}\) is the set of in-neighbors of node u
, and \(\overrightarrow{\mathbf{N}}_{v}^{+}\) is the set of out-neighbors of node v. Noted that node s is the seed node, and $\mathbf{1}(u=s)$ is 1 if u is the seed node s and 0 otherwise.

\begin{equation}
\begin{array}{l}
\mathbf{r}_{u}^{+}=(1-c)\left(\sum_{v \in \overleftarrow{\mathbf{N}}_{u}^{+}} \frac{\mathbf{r}_{v}^{+}}{\left|\overrightarrow{\mathbf{N}}_{v}\right|}+\sum_{v \in \overleftarrow{\mathbf{N}}_{u}^{-}} \frac{\mathbf{r}_{v}^{-}}{\left|\overrightarrow{\mathbf{N}}_{v}\right|}\right)+c \mathbf{1}(u=s) \\
\mathbf{r}_{u}^{-}=(1-c)\left(\sum_{v \in \overleftarrow{\mathbf{N}}_{u}^{-}} \frac{\mathbf{r}_{v}^{+}}{\left|\overrightarrow{\mathbf{N}}_{v}\right|}+\sum_{v \in \overleftarrow{\mathbf{N}}_{u}^{+}} \frac{\mathbf{r}_{v}^{-}}{\left|\overrightarrow{\mathbf{N}}_{v}\right|}\right)
\end{array}
\end{equation}

We use the adjacency matrix \( \mathbf{A} \) of the signed graph $\mathcal{G}$ to vectorize Equation 1. Suppose \( \mathbf{A}_{uv} \) is positive or negative for the signed edge \( (u \rightarrow v) \) and 0 otherwise. \( \mathbf{D} \) is the out-degree diagonal matrix, where \( \mathbf{D}_{ii} = \sum_{j} |\mathbf{A}|_{ij} \). The semi-row normalized matrix is \( \tilde{\mathbf{A}} = \mathbf{D}^{-1} \mathbf{A} \). We decompose \( \tilde{\mathbf{A}} \) into two matrices: a positive semi-row normalized matrix (\( \tilde{\mathbf{A}}_+ \)) and a negative semi-row normalized matrix (\( \tilde{\mathbf{A}}_- \)), such that \( \tilde{\mathbf{A}} = \tilde{\mathbf{A}}_+ - \tilde{\mathbf{A}}_- \).

Based on the above equations, we can vectorize Equation 1 as follows:

\begin{equation}
\begin{aligned}
\mathbf{r}^+ &= (1 - c) \left(\tilde{\mathbf{A}}_+^T \mathbf{r}^+ + \tilde{\mathbf{A}}_-^T \mathbf{r}^-\right) + c\mathbf{q} \\
\mathbf{r}^- &= (1 - c) \left(\tilde{\mathbf{A}}_-^T \mathbf{r}^+ + \tilde{\mathbf{A}}_+^T \mathbf{r}^-\right)
\end{aligned}
\end{equation}

where \( \tilde{\mathbf{A}}_+^T \) and \( \tilde{\mathbf{A}}_-^T \) are the transposed adjacency matrices for positive and negative edges, respectively. \(\mathbf{q}\) is the seed node vector at node \( s \). Initially, set \( \mathbf{r}^+ = \mathbf{q} \), \(\mathbf{r}^- = 0 \), and define \( \mathbf{r}' = \begin{bmatrix}\mathbf{r}^+ \\ \mathbf{r}^- \end{bmatrix} ^T \).

Since the simple balance theory with four types of relationships cannot explain the complex imbalanced relationships, introducing the balance attenuation factors \( \beta \) and \( \gamma \) is beneficial. When a negative walker encounters a negative edge at a node, its sign will switch to positive with probability \( \beta \), or remain negative with probability \( 1 - \beta \). Similarly, when the negative walker encounters a positive edge at node \( m \), its sign will remain negative with probability \( \gamma \), or switch to positive with probability \( 1 - \gamma \). The diffusion model with balance attenuation factors becomes:
\begin{equation}
\begin{aligned}
\mathbf{r}^+ &= (1 - c) \left(\tilde{\mathbf{A}}_+^T \mathbf{r}^+ + \beta \tilde{\mathbf{A}}_-^T \mathbf{r}^- + (1 - \gamma) \tilde{\mathbf{A}}_+^T \mathbf{r}^-\right) + c\mathbf{q} \\
\mathbf{r}^- &= (1 - c) \left(\tilde{\mathbf{A}}_-^T \mathbf{r}^+ + \gamma \tilde{\mathbf{A}}_+^T \mathbf{r}^- + (1 - \beta) \tilde{\mathbf{A}}_-^T \mathbf{r}^-\right)
\end{aligned}
\end{equation}

By repeated iterative computation of \( \mathbf{r}^+ \) and \(  \mathbf{r}^- \) , we can concatenate  \(  \mathbf{r}^+ \) and \(  \mathbf{r}^- \) into \(  \mathbf{r} = \begin{bmatrix}  \mathbf{r}^+ \\  \mathbf{r}^- \end{bmatrix} ^T \). Then by computing the error \( \delta \) between the current iteration result \(  \mathbf{r} \) and the previous iteration result \(  \mathbf{r}' \), where \(\delta = || \mathbf{r}- \mathbf{r}'||\), we update \(  \mathbf{r} \) for the next iteration. The iteration stops when the error \( \delta \) is smaller than a tolerance \(\epsilon\). 

For each node, the algorithm return an \( \mathbf{r}^+ \) and an \( \mathbf{r}^- \). We combine all \( r^+ \) and \( r^- \) into a positive matrix \( \mathbf{r_p} \in \mathbb{R}^{n \times n} \) and a negative matrix \( \mathbf{r_n} \in \mathbb{R}^{n \times n} \). Note that the relationship between edges (\( u \to v \)) and (\( v \to u \)) may differ, and the sign may even be opposite. This means \( \mathbf{r_p}(u,v) \ne \mathbf{r_p}(v,u)\) may occur.

To address this and generate an undirected graph for subsequent process, we transpose \( \mathbf{r_p} \) and \( \mathbf{r_n} \) to \( \mathbf{r_p}^T \) and \( \mathbf{r_n}^T \) respectively where \( \mathbf{r_p}^T(u,v) = \mathbf{r_p}(v,u) \) ( or \( \mathbf{r_n}^T(u,v) = \mathbf{r_n}(v,u) \)). For each node, we take the maximum value between \( \mathbf{r_p} \) and \( \mathbf{r_p}^T \) to form \( \mathbf{r_p}_{\text{max}} \) and \( \mathbf{r_n}_{\text{max}} \). Then, we calculate \( \mathbf{r_p}_{\text{max}} - \mathbf{r_n}_{\text{max}}\) for each node at the corresponding position to generate the symmetric \( \mathbf{r_d} \) matrix, which can also be called the diffuse matrix \(\mathcal{S}\).

\begin{algorithm}
\caption{Diffuse Algorithm}
\label{diffuse}
\begin{algorithmic}[1]
\Require Signed adjacency matrix $\mathbf{A}$
\Ensure Positive score vector: $r^+$ and negative score vector: $r^-$
\State Compute out-degree matrix $\mathbf{D}$ of $|\mathbf{A}|$, $D_{ii} = \sum_{j} |\mathbf{A}|_{ij}$
\State Compute semi-row normalized matrix, $\mathbf{\tilde{A}} = \mathbf{D}^{-1}\mathbf{A}$
\State Split $\mathbf{\tilde{A}}$ into $\mathbf{\tilde{A}}^+$ and $\mathbf{\tilde{A}}^-$ such that $\mathbf{\tilde{A}} = \mathbf{\tilde{A}}^+ - \mathbf{\tilde{A}}^-$

\State Set the starting vector $q$ from the seed node $s$
\State Set $r^+ = q$, $r^- = 0$, and $r' = [r^+; r^-]^\textbf{T}$
\Repeat
\State $r^+ \gets (1 - c)(\mathbf{\tilde{A}}^+ r^+ + \beta \mathbf{\tilde{A}}^- r^- + (1 - \gamma) \mathbf{\tilde{A}}^+ r^-) + cq$
\State $r^- \gets (1 - c)(\mathbf{\tilde{A}}^- r_+ + \gamma \mathbf{\tilde{A}}^+ r^- + (1 - \beta) \mathbf{\tilde{A}}^- r^-)$
\State Concatenate $r^+$ and $r^-$ into $r = [r^+; r^-]$
\State Compute the error between $r$ and $r_0$, $\delta = ||r - r'||$
\State Update $r' \gets r$ for the next iteration
\Until{$\delta < \epsilon$}
\State \textbf{Return:} $r^+$ and $r^-$
\State Initialize empty matrices $rp$ and $rn$ with dimensions $n \times n$
\For{$i$ \textbf{in} range$(1, n)$}
    \For{$j$ \textbf{in} range$(1, n)$}
        \State Set $\mathbf{r_p}(i, j) = \max(r^+(i, j), r^+(j, i))$
        \State Set $\mathbf{r_n}(i, j) = \max(r^-(i, j), r^-(j, i))$
    \EndFor
\EndFor
\State \Return $\mathbf{r_p}, \mathbf{r_n}$
\end{algorithmic}
\end{algorithm}

\subsection{Graph Augmentation}

Generating different views is crucial in contrastive learning methods.
In this study, we primarily focus on randomly removing edges on both the original graph and the diffusion graph to obtain different views.
For each graph, we can construct a matrix \(\mathcal{M} \in \mathbb{R}^{2 \times n} \), where each column represents one edge (such as u->v) containing only existing edges. Then we generate a random masking matrix \(\widetilde{\mathcal{R}} \in \mathbb{R}^{1 \times n}\) drawn from a uniform distribution over $[0,1]$, denoted as \(\widetilde{\mathcal{R}}_{i} \sim \text{Uniform}(0,1)\). Setting a threshold \(p_r =0.1\), we reset \(\widetilde{\mathcal{R}}_{i}=0\) to denote the deletion of the corresponding edge if \(\widetilde{\mathcal{R}}_{i} < 0.1\). The resulting matrix can be computed as: \(\widetilde{\mathcal{M}}=\mathcal{M} \circ \widetilde{\mathcal{R}}\), where \( (x \circ y)_i = x_i y_i \)
is Hadamard product.

We perform this process twice on both the original graph \(\mathcal{G}\) and the diffusion graph \(\mathcal{S}\) separately, randomly removing 10\% of the edges each time. Therefore, we can obtain four augmented graphs, denoted as \(\mathcal{G}_1\), \(\mathcal{G}_2\), \(\mathcal{G}_3\), and \(\mathcal{G}_4\), respectively. 

\subsection{Graph Encoder}
After data augmentation, we obtain four augmented graphs including \(\mathcal{G}_1\), \(\mathcal{G}_2\), \(\mathcal{G}_3\), and \(\mathcal{G}_4\). For convenience, we define the set of augmented graphs as \(\mathbf{G}_k=\left \{ \mathcal{G}_1,\mathcal{G}_2,\mathcal{G}_3,\mathcal{G}_4 \right \}\), where k=1,2,3,4.  Edge types are classified into positive and negative, meaning the two types of effects of genes on specific phenotypes, namely up-regulation and down-regulation. Consequently, it is imperative to design two distinct GNNs to separately aggregate information from positive and negative neighbors. We split each graph into two graphs containing only positive or negative edges, where \(\mathbf{G}_i^{+}=(\mathcal{U} \cup \mathcal{V}, \mathcal{E}^+)\) and \(\mathbf{G}_i^{-}=(\mathcal{U} \cup \mathcal{V}, \mathcal{E}^-)\). Utilizing a design akin to \cite{shu2021sgcl}, node representations are learned from positive graphs using a positive GNN, and from negative graphs using a negative GNN. Parameters are shared in the same perspective for positive (negative) GNNs. GAT model \cite{velivckovic2017graph} is used as the graph encoder and it computes as follows:
\begin{equation}
\begin{aligned}
    h_{i,k}^{(l+1),\zeta} = \text{GAT}_k^\zeta \left(h_{i,k}^{(l),\zeta}, \mathbf{G}_k^\zeta\right)
\end{aligned}
\end{equation}
\begin{equation}
\begin{aligned}
    z_{i,k}^{\zeta} = \left[ h_{i,k}^{(0),\zeta} \parallel h_{i,k}^{(1),\zeta} \parallel \text{···} \parallel h_{i,k}^{(L)} \right] \mathcal{W}_k^{\zeta} 
\end{aligned}
\end{equation}
where $\zeta \in \left \{ +,- \right \}$, $k$ represents the $k$-th augmented graph, $L$ denotes the number of GNN layers and $\mathcal{W}_k^{\zeta}$ is a learnable transformation matrix. $h_{i,k}^{(0),\zeta}$ denotes the input feature vector of the $i$-th node and $h_{i,k}^{(l),\zeta}$ is the representation of the $i$-th node for the $l$-th layer. $z_{i,k}^{\zeta}$ represents the final representation of $i$-th node in the $k$-th augmented graph.
Note that we use two GAT layers here.

\subsection{Objective Loss}
\subsubsection{Contrastive Loss}

\textbf{Inter-view Contrastive Learning.}
As mentioned earlier, we obtained four augmented graphs and further divided each augmented graph into positive and negative graphs for separate encoding. Since the positive and negative graphs contain different semantic properties, we define inter-view contrastive losses for both the positive and negative augmented graphs. Below, we discuss the positive graphs in detail and define the losses for the negative graphs in a similar manner.

For any positive graph, in order to obtain more robust representations, CSGDN maximizes the agreements of the representations between the same node across different positive graphs while minimizing the representation similarities between different nodes. For example, the representation of the \(i\)-th node in graph \( \mathbf{G}_1^{+}\) of the inter-set perspective, i.e., \(z_{i, 1}^+\), should be consistent with representations generated from the same node in the other positive augmented graph \( \mathbf{G}_2^{+}\) in the same perspective, i.e., \(z_{i, 2}^+\). Therefore, we treat the representations of the same node from other positive graphs within the same perspective as positive samples. Also, we want the representation of a node to be distinct from those of different nodes so we consider the representations of different nodes from other positive graphs within the same perspective as negative samples. 
Given a mini-batch \(\mathcal{B}\) containing I nodes, the inter-view contrastive loss for positive augmented graphs is defined as follows, inspired by the InfoNCE loss \cite{shu2021sgcl,sohn2016improved}:

\begin{equation}    
\begin{aligned}
    \mathcal{L}_{\text{inter}}^{+}=-\frac{1}{I} \sum_{i=1}^{I} \log \frac{\exp \left(\frac{\operatorname{sim}\left(z_{i, k}^{+}, z_{i, k^{\prime}}^{+}\right)}{\tau}\right)}{\sum_{j=1, j \neq i}^{I} \exp \left(\frac{\operatorname{sim}\left(z_{i, k}^{+}, z_{j, k^{\prime}}^{+}\right)}{\tau}\right)}
\end{aligned}
\end{equation}

where \(z_{i, k}^{+}\) represents the representation of node \(i\) in the \(k\)-th augmented positive graph, sim(·,·) represents the similarity function between the two representations and \(\tau\) denotes the preset temperature parameter.
 
Similarly, for the representation of the \(i\)-th node in the graph \( \mathbf{G}_1^{-} \) of the inter-set perspective \(z_{i,1}^-\), its inter-view positive samples are the  representations generated from the same node from other negative graphs , and its inter-view negative samples are the ones generated from different nodes in other negative graphs. As the same, the inter-view contrastive loss for negative augmented graphs is defined as:

\begin{equation}    
\begin{aligned}
    \mathcal{L}_{\text{inter}}^{-} = -\frac{1}{I} \sum_{i=1}^{I} \log \frac{\exp \left( \frac{\operatorname{sim}\left(z_{i, k}^{-}, z_{i, k^{\prime}}^{-}\right)}{\tau} \right)}{\sum_{j=1, j \neq i}^{I} \exp \left( \frac{\operatorname{sim}\left(z_{i, k}^{-}, z_{j, k^{\prime}}^{-}\right)}{\tau} \right)}
\end{aligned}
\end{equation}

Combining the above two losses, we obtain the perspective specific contrastive loss:
\begin{equation}    
\begin{aligned}
    \mathcal{L}_{\text{inter}}=\mathcal{L}_{\text{inter}}^{+} + \mathcal{L}_{\text{inter}}^{-}
\end{aligned}
\end{equation}

\textbf{Intra-view Contrastive Learning.}
In addition to maximizing the consistency of representations for the same node across different positive graphs or different negative graphs, we also design intra-view contrastive losses to let the ultimate representation of each node to be close to the representations of the same node in positive graphs and far from the representations of the same node in negative graphs.

For node \(v_i\), we generate the representation by concatenating all representations containing different information of diverse views, which is formulated as follows:
\begin{equation}
\begin{aligned}
    z_{i} = g \left(z_{i,1}^{+} \parallel z_{i,2}^{+} \parallel z_{i,3}^{+} \parallel z_{i,4}^{+} \parallel z_{i,1}^{-} \parallel z_{i,2}^{-} \parallel z_{i,3}^{-} \parallel z_{i,4}^{-} \right) ,
\end{aligned}
\end{equation}
where \(g\) is a two-layer MLP, and \(z_{i} \in \mathbb{R}_\text{d}\) represents the final representation of node \(v_i\).

To be specific, we treat the representations
of the same node from other positive graphs   as positive samples while the representations of the same node from other negative graphs as negative samples.
Given a mini-batch $\mathcal{B}$ containing I nodes, the intra-view contrastive objective is formally defined as:

\begin{equation}
\begin{aligned}
\mathcal{L}_{\text{intra}} = -\frac{1}{I} \sum_{i=1}^{I} \log \left( \frac{\sum_{m=1}^{M} \exp\left( \frac{\text{sim}(z_i, z_{i,m}^{+})}{\tau} \right)}{\sum_{m=1}^{M} \exp\left( \frac{\text{sim}(z_i, z_{i,m}^{-})}{\tau} \right)} \right),
\end{aligned}
\end{equation}
where \(M\) denotes the number of graph views, which equals to 4 in this paper.

\textbf{Contrastive loss.} 
we generates the combined contrastive learning objective from the inter-view and intra-view contrastive learning objectives, and it is formulated as follows:

\begin{equation}
\begin{aligned}
    \mathcal{L}_\text{CL} = (1-\alpha)\mathcal{L}_\text{inter} + \alpha\mathcal{L}_\text{intra},
\end{aligned}
\end{equation}
where \(\alpha\) is the weight coefficient that controls the significance between two losses.

\subsubsection{Label Loss}

After obtaining the final node representations using Equation 9, we utilize a two-layer MLP to compute the sign scores between nodes from different sets:

\begin{equation}
\begin{aligned}
    \hat{y}_{\text{pred}} = \text{sigmoid}(\text{MLP}(z_{u_i} \parallel z_{v_j})),
\end{aligned}
\end{equation}

where \(\hat{y}_{\text{pred}}\) represents the predicted sign score of the link between nodes \(u_i \in U\) and \(v_j \in V\). The dimension of \(\hat{y}_{\text{pred}}\) is 3, representing the probabilities of predicting the edge sign as positive, negative, or neutral, respectively.

Following existing methodologies, the cross-entropy loss function is used for link sign prediction:

\begin{equation}
\begin{aligned}
    \mathcal{L}_{\text{label}} = -\sum_{i=1}^{N}  y \log(\hat{y}_{\text{pred}})
\end{aligned}
\end{equation}

where  $y$ is the ground truth label. The labels are defined as follows:
\begin{itemize}
    \item -1 indicates a negative relationship (down-regulation).
    \item 1 indicates a positive relationship (up-regulation).
    \item 0 indicates an undefined relationship which is not from the TWAS analysis process.
\end{itemize}
Note that the ground truth label are converted into one-hot encoding, with a value of 1 in the corresponding category position and 0 elsewhere.

\subsubsection{Total Loss}
Finally, our model is trained using a joint loss function that integrates the link sign prediction loss and the contrastive learning loss:

\begin{equation}
\begin{aligned}
\mathcal{L} = \mathcal{L}_{\text{label}} + \beta \mathcal{L}_{\text{CL}},
\end{aligned}
\end{equation}

where \(\beta\) is a weight parameter that controls the relative importance of the contrastive loss.

\begin{algorithm}
\caption{Contrastive Algorithm}
\label{contrastive}
\begin{algorithmic}[1]
\For{$epoch = 0, 1, \ldots$}
    \State // Graph Augmentations
    \State Generate four graph views $\mathbf{G}_k$ by perturbing $\mathcal{G}$ and $\mathcal{S}$
    \State // Graph Encoders
    \State Split $\mathbf{G}_k$ into $\mathbf{G}_{k}^+$, $\mathbf{G}_{k}^-$
    \State Obtain consistent representations of $\mathbf{G}_{k}^+$ 
    \State Obtain inconsistent representations of $\mathbf{G}_{k}^-$
    \State Obtain the ultimate representations $Z$
    \State // Contrastive Learning
    \State Compute inter-view contrastive loss $\mathcal{L}_{inter}$
    \State Compute intra-view contrastive loss $\mathcal{L}_{intra}$
    \State Compute the combined contrastive objective $\mathcal{L}_{CL}$
    \State // Model Training
    \State Compute the loss of sign link prediction task
    \State Compute the whole objective function $\mathcal{L}$ via and update model parameters $\theta$ by $\frac{\partial}{\partial \theta} \mathcal{L}$
\EndFor
\State \textbf{return} node representations $Z$
\end{algorithmic}
\end{algorithm}

\subsection{MLP for Genes without TWAS associations}

\begin{figure}[ht]
    \begin{minipage}{0.45\textwidth}
    \centering
    \includegraphics[width=\textwidth]{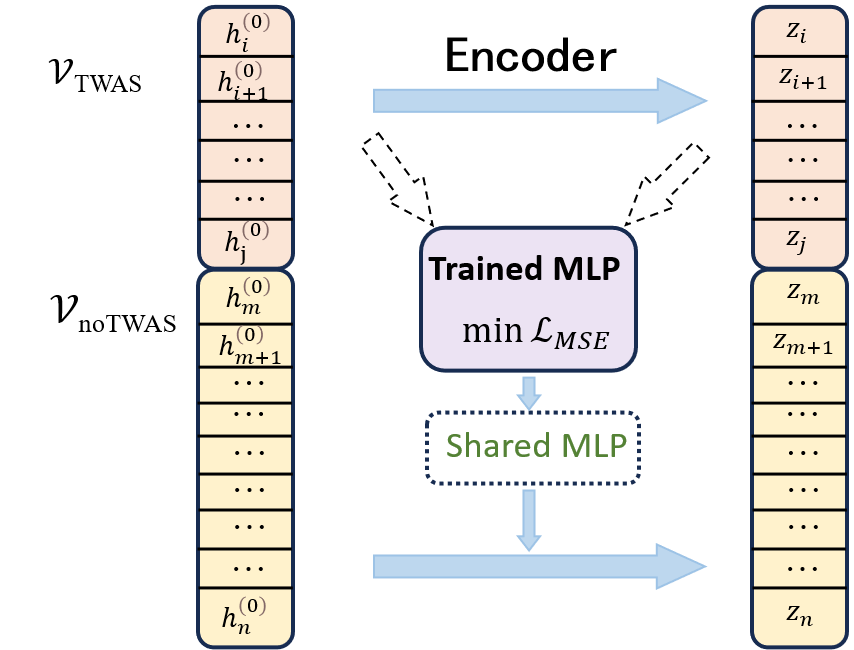}
    \caption{The frame for genes can not be associated with phenotypes.}
    \label{fig:MLP}
    \end{minipage}
\end{figure} 

For those genes that lack TWAS associations, due to the lack of supervision from TWAS association information, the encoder trained specifically for genes in TWAS associations cannot be directly used for their encoding. 
To address this issue, we propose to train a multi-layer perceptron (MLP) to transform these genes that lack TWAS associations into a shared space with TWAS-associated genes:

\begin{equation}
\mathcal{L}_{\text{MSE}} = \frac{1}{|\mathcal{V}_\text{TWAS}|} \sum_{v_i \in \mathcal{V}_\text{TWAS}} \left\| \text{MLP}(h_{i}^{(0)})  - z_i \right\|^2
\end{equation}

where $h_{i}^{(0)}$ represents the input feature of gene $v_i$ in TWAS associations, and $z_i$ represents the final representation of gene $v_i$ after being encoded by the aforementioned TWAS framework. By minimizing the above Mean Squared Error (MSE) loss, we can learn the encoding capability of the main framework for TWAS-associated genes through MLP, which can be used for the encoding of genes lacking TWAS associations. Then we can obtain the representation of genes lacking TWAS associations through this MLP:
\begin{equation}
z_i = \text{MLP}(h_{i}^{(0)})
\end{equation}
where $v_i \in \mathcal{V}_\text{noTWAS}$. The specific process is shown in Fig. 3.

\section{EXPERIMENTS}

\begin{table*}[]
\centering
\caption{Link sign prediction results (average standard deviation) with AUC, Binary-F1, Micro-F1 and Macro-F1 (\%) on 2 benchmark datasets.}
\tiny
\resizebox{\textwidth}{!}{
\begin{tabular}{c c|c c c|c c c c}
\hline
\multirow{2}{*}{Datasets} &  & \multicolumn{3}{c|}{Unsigned GNNs} & \multicolumn{4}{c}{Signed GNNs} \\
\cline{3-9} 
&  & GCN & GAT & GRACE & SGCN & SGCL & SGNNMD & Our CSGDN \\
\hline
\multirow{4}{*}{\textit{G.\ hirsutum}} & AUC & 0.7145 ± 0.0294 & 0.7098 ± 0.0093 & 0.7085 ± 0.0068 & 0.7215 ± 0.0123 & 0.7282 ± 0.0191 & 0.6883 ± 0.1783 & \textbf{0.7811 ± 0.0116} \\
 & Binary-F1 & 0.6231 ± 0.0633  & 0.5950 ± 0.0131 & 0.5934 ± 0.0098 & 0.6325 ± 0.0190  & 0.6307 ± 0.0365 & \textbf{0.7894 ± 0.1128} & 0.7458 ± 0.0103 \\
 &Micro-F1 & 0.7444 ± 0.0082  & 0.7624 ± 0.0102 & 0.7609 ± 0.0074 & 0.7624 ± 0.0122  & 0.7759 ± 0.0146 & 0.7203 ± 0.1578 & \textbf{0.7804 ± 0.0240} \\
 &Macro-F1 & 0.7102 ± 0.0139 & 0.7134 ± 0.0104 & 0.7120 ± 0.0076 & 0.7284 ± 0.0136 & 0.7349 ± 0.0221 & 0.6634 ± 0.2151 & \textbf{0.7981 ± 0.0447} \\
 \hline
 \multirow{4}{*}{\textit{B.\ napus}} & AUC & 0.5000 ± 0.0000 & 0.5000 ± 0.0000 & 0.5000 ± 0.0000 & 0.5821 ± 0.0449 & 0.4409 ± 0.1262 & 0.6026 ± 0.1397 & \textbf{0.6615 ± 0.0495} \\
 & Binary-F1 & \textbf{0.9130 ± 0.0000} & \textbf{0.9130 ± 0.0000} & 0.8773 ± 0.0000 & 0.8560 ± 0.0385 & 0.5133 ± 0.3839 & 0.2456 ± 0.3027 & 0.8608 ± 0.0334 \\
& Micro-F1 & 0.8400 ± 0.0000 & 0.8400 ± 0.0000 & 0.7815 ± 0.0000 & 0.7627 ± 0.0574 & 0.4800 ± 0.2918 & \textbf{0.8233 ± 0.0304} & 0.7815 ± 0.0437 \\ 
& Macro-F1 & 0.4565 ± 0.0000 & 0.4565 ± 0.0000 & 0.4387 ± 0.0000 & 0.5828 ± 0.0550 & 0.3368 ± 0.1848 & 0.5719 ± 0.1558 & \textbf{0.6649 ± 0.0478} \\ \hline
\multirow{4}{*}{\textit{T.\ turgidum}} & AUC & 0.5000±0.0000 & 0.5000±0.0000 & 0.5000±0.0000 & 0.4714±0.0350 & 0.5000±0.0000& 0.5810±0.1160
& \textbf{0.5982+0.0787} \\
&Binary-F1&0.8235±0.0000&0.8235±0.0000&\textbf{0.8636±0.0000}&0.7941±0.0360& 0.8235±0.0000&0.3352±0.2201&0.8231+0.0946 \\
&Micro-F1&0.7000±0.0000&0.7000±0.0000&0.7600±0.0000&0.6600±0.0490&0.7000±0.0000& \textbf{0.7886±0.1022}&0.5925+0.1027 \\
&Macro-F1&0.4118±0.0000&0.4118±0.0000&0.4318±0.0000&0.3971±0.0180&0.4118±0.0000&0.6043±0.1411&\textbf{0.5925+0.1027}\\
\hline

\end{tabular}
}
\end{table*}

\begin{table}[]
\centering
\caption{Link sign prediction results (average standard deviation) with AUC, Binary-F1, Micro-F1, Macro-F1 (\%) for randomly sampled 80\% \textit{G.\ hirsutum} dataset.}
\tiny
\resizebox{0.47\textwidth}{!}{
\begin{tabular}{c|c c c c}
\hline
&AUC & Binary-F1 & Micro-F1 & Macro-F1 \\
\hline
GCN & 0.5883 ± 0.0182 &0.4084 ± 0.0932& 0.6562 ± 0.0393 &0.5792 ± 0.0317 \\
GAT & 0.5549 ± 0.0244 &0.3042 ± 0.1045&0.6315 ± 0.1417&0.4811 ± 0.1058\\
GRACE & 0.6967 ± 0.0208&0.6203 ± 0.0330&0.7098 ± 0.0504&0.6849 ± 0.0378\\
\hline
SGCN & 0.5832 ± 0.0478&0.4134 ± 0.0885&0.6472 ± 0.0208&0.5793 ± 0.0431 \\
SGCL&0.6271 ± 0.0429&0.4980 ± 0.0975&0.6360 ± 0.0464&0.5926 ± 0.0174 \\
SGNNMD&0.5988 ± 0.1210&0.5035 ± 0.2912&0.5964 ± 0.1242&0.5231 ± 0.1843 \\
\hline
CSGDN&\textbf{0.7495 ± 0.0339}&\textbf{0.6884 ± 0.0407}&\textbf{0.7574 ± 0.0515}&\textbf{0.7406 ± 0.0420}\\
\hline

\end{tabular}
}
\end{table}

\begin{table}[]
\centering
\caption{Link sign prediction results (average standard deviation) with AUC, Binary-F1, Micro-F1, Macro-F1 (\%) for the \textit{G.\ hirsutum} dataset with perturbations.}
\tiny
\resizebox{0.47\textwidth}{!}{
\begin{tabular}{c|c|c c c|c c c c}
\hline
  &\multirow{2}{*}{Ptb(\%)} &  \multicolumn{3}{c|}{Unsigned GNNs} & \multicolumn{4}{c}{Signed GNNs} \\
\cline{3-9} 
 & & GCN & GAT & GRACE & SGCN & SGCL & SGNNMD & CSGDN \\
\hline
AUC&\multirow{4}{*}{10}&0.6714±0.0116&0.6649±0.0169&0.6759±0.0163&0.6994±0.0138&0.6850±0.0221&0.6258±0.1102&\textbf{0.7066+0.0300}\\
Binary-F1&&0.5424±0.0302&0.5166±0.0387&0.5435±0.0309&0.6046±0.0330&0.5663±0.0429&0.5547±0.2151&\textbf{0.6498+0.0562}\\
Micro-F1&&0.7203±0.0129&0.7203±0.0129&0.7278±0.0138& \textbf{0.7368±0.0106}&0.7323±0.0169&0.6045±0.1409&0.6987+0.0253\\
Macro-F1&&0.6702±0.0150&0.6598±0.0220&0.6748±0.0196&\textbf{0.7033±0.0159}&0.6863±0.0259&0.5891±0.1427&0.6987+0.0253\\
\hline
AUC&\multirow{4}{*}{20}&0.6030 ± 0.0454&0.6313 ± 0.0176&0.6515 ± 0.0061&0.6225 ± 0.0312&0.6518 ± 0.0198&0.6063 ± 0.1025&\textbf{0.7209 ± 0.0298}\\
Binary-F1&&0.4939 ± 0.0901&0.4863 ± 0.0317&0.5320 ± 0.0047&0.5576 ± 0.0403&0.5840 ± 0.0338&0.6420 ± 0.1803&\textbf{0.6501 ± 0.0530}\\
Micro-F1&&0.6286 ± 0.0400&0.6797 ± 0.0162&0.6932 ± 0.0111&0.6316 ± 0.0293&0.6602 ± 0.0573&0.6180 ± 0.1165&\textbf{0.7444 ± 0.0190}\\
Macro-F1&&0.5968 ± 0.0468&0.6268 ± 0.0207&0.6517 ± 0.0053&0.6205 ± 0.0306&0.6364 ± 0.0408&0.5847 ± 0.1244&\textbf{0.7236 ± 0.0289}\\
\hline
\end{tabular}
}
\end{table}

\begin{table}[]
\centering
\caption{The AUC performances with CSGDN and its variants}
\tiny
\resizebox{0.47\textwidth}{!}{
\begin{tabular}{c|cccc}
\hline
& CSGDN & $\text{CSGDN}_{\mathrm{w/o\ diffuse}}$ & $\text{CSGDN}_{\mathrm{w/o\ aug}}$ & $\text{CSGDN}_{\mathrm{w/o\mathcal{L}_{CL}}}$ \\
\hline
\textit{G.\ hirsutum} & \textbf{0.7811 ± 0.0116}&0.7624 ± 0.0179&0.7712 ± 0.0551&0.7297 ± 0.0208\\
\textit{B.\ napus} & 0.6615 ± 0.0495&0.6485 ± 0.0400&\textbf{0.6700 ± 0.0206}&0.6204 ± 0.0017\\
\textit{T.\ turgidum} & \textbf{0.5982 ± 0.0787} &0.5175 ± 0.0351 &0.5333 ± 0.0887&0.4956 ± 0.0428\\
80\% \textit{G.\ hirsutum} & \textbf{0.7495 ± 0.0339}&0.6998 ± 0.0365&0.7355 ± 0.0184&0.6870 ± 0.0327\\
\hline
\end{tabular}
}
\end{table}

\begin{table*}[]
\centering
\caption{10 genes associated with \textit{G.\ hirsutum} phenotypes and their types predicted by CSGDN.}
\tiny
\resizebox{\textwidth}{!}{
\begin{tabular}{cccccc|cccccc|cccccc|cccccc}
    \toprule
    \multicolumn{6}{c|}{FE} & \multicolumn{6}{c|}{FU} & \multicolumn{6}{c|}{FS} & \multicolumn{6}{c}{FL} \\
    \textbf{Gene} & \textbf{TWAS} & \textbf{CSGDN} & \textbf{$P_{\text{down}}$} & \textbf{$P_{\text{none}}$} & \textbf{$P_{\text{up}}$} & \textbf{Gene} & \textbf{TWAS} & \textbf{CSGDN} & \textbf{$P_{\text{down}}$} & \textbf{$P_{\text{none}}$} & \textbf{$P_{\text{up}}$} & \textbf{Gene} & \textbf{TWAS} & \textbf{CSGDN} & \textbf{$P_{\text{down}}$} & \textbf{$P_{\text{none}}$} & \textbf{$P_{\text{up}}$} & \textbf{Gene} & \textbf{TWAS} & \textbf{CSGDN} & \textbf{$P_{\text{down}}$} & \textbf{$P_{\text{none}}$} & \textbf{$P_{\text{up}}$} \\
    \midrule
    Ghir\_A13G012290 & Up & Down & 0.672 & 0.033 & 0.294 & Ghir\_A02G005400 & Down & Up & 0.043 & 0.013 & 0.944 & Ghir\_D09G001870 & Up & Up & 0.014 & 0.007 & 0.979 & Ghir\_A09G017330 & Up & Up & 0.010 & 0.006 & 0.984 \\ 
    Ghir\_D02G002560 & Up & Up & 0.012 & 0.007 & 0.982 & Ghir\_A05G041310 & Up & Up & 0.014 & 0.007 & 0.979 & Ghir\_D07G018780 & Up & Up & 0.089 & 0.019 & 0.892 & Ghir\_D07G002590 & Up & Up & 0.010 & 0.006 & 0.984 \\ 
    Ghir\_A11G026810 & Up & Up & 0.012 & 0.007 & 0.981 & Ghir\_A04G002650 & Up & Up & 0.031 & 0.011 & 0.958 & Ghir\_D05G001070 & Up & Up & 0.012 & 0.007 & 0.982 & Ghir\_D01G006880 & Down & Up & 0.282 & 0.031 & 0.686\\ 
    Ghir\_A09G016050 & Up & Up & 0.452 & 0.035 & 0.513 & Ghir\_D03G012470 & Up & Up & 0.012 & 0.007 & 0.981 & Ghir\_D07G018070 & Down & Up & 0.018 & 0.008 & 0.973 & Ghir\_D05G006550 & Down & Down & 0.990 & 0.006 & 0.004 \\ 
    Ghir\_A12G025670 & Up & Down & 0.483 & 0.035 & 0.482 & Ghir\_A12G019480 & Down & Down & 0.990 & 0.006 & 0.004 & Ghir\_D07G002590 & Down & Down & 0.493 & 0.035 & 0.472 & Ghir\_A01G012860 & Down & Down & 0.589 & 0.035 & 0.376 \\ 
    Ghir\_A09G014910 & Down & Down & 0.971 & 0.011 & 0.018 & Ghir\_A12G019760 & Down & Up & 0.019 & 0.009 & 0.972 & Ghir\_D06G018900 & Down & Down & 0.991 & 0.005 & 0.004 & Ghir\_A01G017490 & Down & Down & 0.981 & 0.009 & 0.011 \\ 
    Ghir\_A01G013950 & Down & Down & 0.989 & 0.006 & 0.005 & Ghir\_A01G018190 & Down & Down & 0.990 & 0.006 & 0.004 & Ghir\_D13G017280 & Down & Down & 0.991 & 0.005 & 0.004 & Ghir\_A01G013320 & Down & Down & 0.987 & 0.007 & 0.006 \\ 
    Ghir\_D01G000640 & Down & Down & 0.928 & 0.018 & 0.054 & Ghir\_A01G017780 & Down & Down & 0.991 & 0.005 & 0.004 & Ghir\_A01G016960 & Down & Down & 0.989 & 0.006 & 0.005 & Ghir\_A11G001240 & Down & Up & 0.038 & 0.012 & 0.949 \\ 
    \bottomrule
\end{tabular}
}
\end{table*}

In this section, we present experiments on real-world datasets to evaluate the effectiveness of CSGDN in link sign prediction. We also compare its performance with leading methods in both unsigned and signed graph neural networks. Specifically, we aim to address the following questions:

\begin{itemize}
    \item \textbf{Q1}: Does CSGDN outperform the advanced baselines?
    \item \textbf{Q2}: How does CSGDN perform with a small sample size and random noise?
    \item \textbf{Q3}: How do different model components affect the performance of CSGDN?
    \item \textbf{Q4}: Is CSGDN sensitive to hyperparameters?
\end{itemize}

\subsection{Experiment Settings}

\textbf{Baselines.} 
To validate the effectiveness of CSGDN, we compare our proposed model with several common methods in the fields of unsigned and signed graph neural networks.

$\bullet$ Unsigned GNNs. GCN \cite{kipf2016semi} is a pioneering and notable GNN model tailored for unsigned graphs, featuring an effective layer-wise propagation mechanism. GAT \cite{velivckovic2017graph} utilizes masked self-attentional layers, allowing nodes to attend to their neighbors’ features with varying weights without costly matrix operations or prior knowledge of the graph structure. GRACE \cite{zhu2020deep} proposes a novel unsupervised graph representation learning framework that leverages contrastive objectives at the node level, creating two graph views through corruption, and maximizing the agreement of node representations in these views, utilizing diverse contexts via a hybrid scheme on structure and attribute levels, and demonstrating superior performance over state-of-the-art methods.

$\bullet$ Signed GNNs. SGCN \cite{derr2018signed} utilizes balance theory to correctly aggregate and propagate the information across layers of a signed GCN model and generalizes GCN to signed graphs. SGCL \cite{shu2021sgcl} introduces a novel graph contrastive representation learning techniques tailored for signed graphs, leveraging balance theory and dual contrastive strategies to achieve superior node representations across diverse datasets, including social and online gaming networks. SGNNMD \cite{zhang2022sgnnmd} utilizes signed graph neural networks to predict deregulation types of miRNA-disease associations, achieving competitive performance by integrating structural and biological features from a signed bipartite network.

\textbf{Hyper-parameters setting.} 
we analyze the sensitivity of CSGDN to six key hyperparameters: $\alpha$, $\beta$, $feature$, $mask$, $predictor$, and $\tau$. The default configuration for each hyperparameter is as follows: $\alpha = 0.8$, $\beta = 0.01$, $feature = 64$, $mask = 0.4$, $predictor = \text{2-layer MLP}$, and $\tau = 0.05$. These settings were determined based on the model’s overall highest AUC score. The AUC value is utilized as the primary metric to assess the sensitivity of the model to changes in these hyperparameters.

\textbf{Reducing sample size.} 
We randomly extract eighty percent of the \textit{G.\ hirsutum} dataset. CSGDN shows the excellent results to face with small sample size. 

\textbf{Random Noise.} 
To demonstrate that our model CSGDN has an outstanding performance when resisting interference,  we achieve the effect of random perturbation to simulate noise by randomly flipping  a certain proportion of edge signs in the \textit{G.\ hirsutum} dataset. In this experiment, we set the proportion as 10\% and 20\%.

\textbf{Task and evaluation metrics.} 
We use AUC, Micro-F1,Binary-F1 and Macro-F1 to evaluate the results on the link sign prediction task. For each dataset, we randomly split edges into a training set and a testing set with a ratio 8:2. Note that superior performance is indicated by higher values for all these four evaluation metrics.

\subsection{Experiment Results}

\textbf{Performance of CSGDN compared with baselines (Q1).} To answer \textbf{Q1}, we compare CSGDN with current state-of-the-art methods. We primarily use two types of GNN frameworks as baselines: unsigned GNN and signed GNN. For unsigned GNNs, we used GCN, GAT, and GRACE, while for signed GNNs, we employed SGCN, SGCL, and SGNNMD. We use link prediction as the evaluation task, with AUC, Binary-F1, Micro-F1, and Macro-F1 as the evaluation metrics for model performance. Across three common crop datasets (\textit{G.\ hirsutum}, \textit{B.\ napus}, and \textit{T.\ turgidum}), the evaluation metrics of CSGDN generally outperform those of the state-of-the-art baselines. As shown in Table 2, CSGDN demonstrates strong performance in the link prediction task on crop datasets.

\textbf{Performance of CSGDN when addressing small sample size and random noise (Q2).} As shown in Tables 3 and 4, we address \textbf{Q2} by verifying that CSGDN can effectively overcome two issues: the costs of samples and noise. For the first issues we randomly reduce the sample size of the \textit{G.\ hirsutum} dataset to 80\% and subsequently divide this dataset into training and testing sets. The results presented in the Table 3 indicate that CSGDN outperforms all baselines on \textit{G.\ hirsutum} datasets with randomly reduced sample sizes, demonstrating its effectiveness in handling small sample datasets. This suggests that we can reduce experimental costs and durations by minimizing the sample size while still achieving excellent prediction outcomes. Then for noise, we utilize two \textit{G.\ hirsutum} datasets with the proportion of 10\% and 20\% perturbations. As shown in Table 4, our model CSGDN outperforms the vast majority of baselines. This reflects that our model has strong anti-interference ability against various types of noise through the contrastive learning methods.

\subsection{Ablation Study}

We conduct ablation study to assess the effectiveness of different components in our proposed model to answer \textbf{Q3}. In this subsection, we employ sign perturbation as the graph augmentation method to analyze performance. Specifically, we compare CSGDN with its three variants: \(\text{CSGDN}_{\mathrm{w/o\ diffuse}}\), \(\text{CSGDN}_{\mathrm{w/o\ aug}}\), and \(\text{CSGDN}_{\mathrm{w/o\mathcal{L}_{CL}}}\), which are defined as follows:

\begin{itemize}
\item \(\text{CSGDN}_{\mathrm{w/o\ diffuse}}\) : The graph diffusion in contrastive learning is removed, and instead, we utilize the same two original graphs without diffusion step for the next augmentation step.

\item \(\text{CSGDN}_{\mathrm{w/o\ aug}}\) : The graph augmentation step is removed. In this variant, graphs $\mathcal{G}$ and $\mathcal{S}$ instead of augmented graphs $\mathbf{G_K}$ are exploited during training.

\item\(\text{CSGDN}_{\mathrm{w/o\mathcal{L}_{CL}}}\) : The part of contrastive learning is removed and ignores the contrastive loss.

\end{itemize}

As shown in Table 5, we demonstrate that all three components are essential to CSGDN’s performance. Each components plays a unique role in enhancing the model’s ability.

\subsection{Hyper-parameters Analysis}

\begin{figure}[ht]  
    \centering
    \begin{minipage}{0.49\textwidth}
        \centering
        \includegraphics[width=\textwidth]{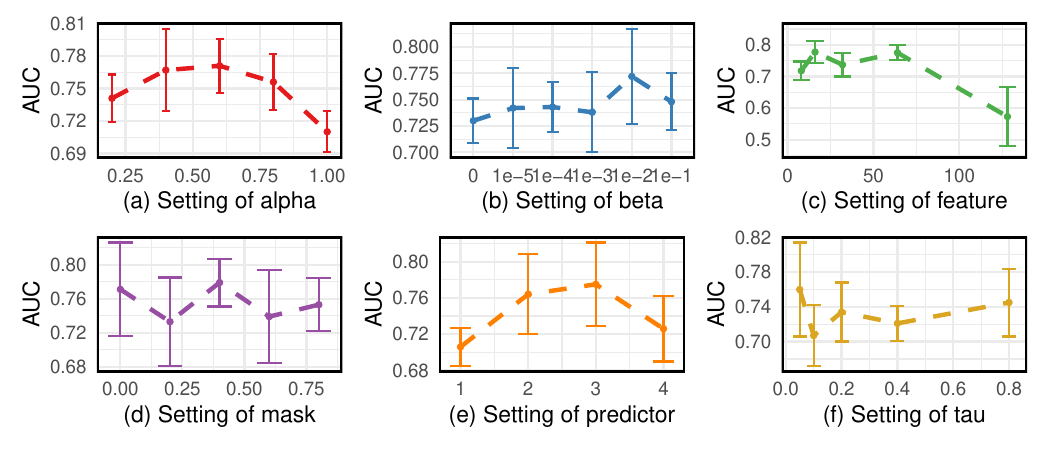}
        \caption{\centering Hyperparameter sensitivity CSGDN in the \textit{G.\ hirsutum} dataset.}
        \label{fig: para}
    \end{minipage}
\end{figure}

To answer \textbf{Q4}, we analyze the sensitivity of CSGDN to six key hyperparameters: $\alpha$, $\beta$, $feature$, $mask$, $predictor$, and $\tau$. 

The hyperparameter $\alpha$, shown in Fig. 4(a), serves as the weight coefficient that balances the significance of the inter-view contrastive loss and intra-view contrastive loss. We evaluate the model’s performance with $\alpha$ set to values from the set ${0.2, 0.4, 0.6, 0.8, 1.0}$. We find that the model performs better when $\alpha$ is between 0.5 and 0.8, while anything outside of this range decreases performance. Therefore, for this dataset intra-view, i.e., learning a consistent representation across augmented graphs, is more important.

The hyperparameter $\beta$, shown in Fig. 4(b), controls the trade-off in the joint loss function between the link sign prediction loss ($\mathcal{L}_{label}$) and the contrastive learning loss ($\mathcal{L}_{CL}$). We vary $\beta$ over the set ${0, 0.1, 0.01, 0.001, 0.0001, 0.00001}$ and observe that the highest AUC score is achieved when $\beta = 0.01$. We find that the model performance tends to increase when $\beta > 0.01$, and the model reaches its best performance at $\beta = 0.01$, followed by a decrease in performance. This illustrates the importance of contrast learning in the model, which decreases when the contrast learning effect on model performance is extreme, i.e., when the $\beta$ value is too small or too large.

The hyperparameter $feature$, shown in Fig. 4(c), refers to the node embedding dimension. We test six different node embedding dimensions, ranging from 8 to 128, and evaluate their impact on model performance. We find that the model performance decreases extremely when $feature = 128$, which may be due to feature sparsity caused by the elevated dimension of the feature space, which prevents the model from effectively establishing correlations between nodes.

The hyperparameter $mask$, shown in Fig. 4(d), represents the ratio of edges randomly dropped from both the original and diffusion graphs. We explore the values of $mask$ in the set ${0, 0.2, 0.4, 0.6, 0.8}$ and find that the model performs optimally when the mask ratio is set to 0.4. We find that the model performance fluctuates as mask ratio increases, but generally performs better when mask ratio is small. When mask ratio is large, too much information is lost, resulting in incomplete information and lower model performance.

The hyperparameter $predictor$, shown in Fig. 4(e), determines the best architechture of the downstream link prediction task, allowing for variations among a 1-layer MLP, 2-layer MLP, 3-layer MLP, and 4-layer MLP. The number of layers directly influences the model’s prediction accuracy. We found that as the number of layers of $predictor$ increases, the model performance first increases and then decreases, and reaches the highest value at 3-layer MLP, while 4-layer MLP may be due to the overfitting of the model because of the excessive number of layers of the neural network, which is ultimately manifested in the decrease of the model performance on the test set.

Finally, the hyperparameter $\tau$, shown in Fig. 4(f), is the temperature parameter used in the contrastive loss function to regulate the similarity contrast between positive and negative edges. We test $\tau$ values from the set ${0.05, 0.1, 0.2, 0.4, 0.8}$, and the default value of $\tau = 0.05$ yields the best performance. We find that the model performance is better when the $\tau$ value is lower, and generally lower when $\tau > 0.05$, which also indicates that this dataset needs to learn a consistent representation as much as possible in the comparative learning, and smoothing operation here will reduce the performance.

\subsection{Case Study}

In this section, we initiate a case study focusing on \textit{G.\ hirsutum} genes that associated with the four types of phenotypes including Fiber Elongation rate (FE), Fiber Uniformity (FU), Fiber Strength (FS), Fiber Length (FL). By leveraging the optimal hyperparameter
configurations outlined,  we utilize
the associations between \textit{G.\ hirsutum} genes and four types phenotypes predicted by TWAS process as the training set to train the CSGDN's TWAS Frame. Then, the randomly-selected gene-phenotype associations types that TWAS cannot calculate in \textit{G.\ hirsutum} are used as irrelevant associations. CSGDN can predict probabilities of three types associations, including up-regulation, down-regulation and irrelevant associations. We take the type of the highest predicted probability as the current gene and phenotype type. Thus, for the target four phenotypes, we can predict association types with target \textit{G.\ hirsutum} genes. As shown in Table 6, we randomly list 10 associations of each phenotype. We use the result associations from TWAS as the reference for our predictions. For example, for phenotype FE, 2 out of 16 predictions are incorrectly compared with TWAS results. For phenotype FU, 2 out of 9 predictions are incorrect from the TWAS results.  Therefore, the case studies demonstrate the usefulness of CSGDN in discovering novel gene-phenotype associations and can be validated by TWAS results.

\section{CONCLUSION}

Association prediction between gene and phenotype plays a crucial role in grasping complex biological and genetic process in crops. We propose a novel model CSGDN to address two major issues including costs and noise. 
CSGDN employs the diffusion method to capture the potential associations with minimal sample size. Contrastive learning strategies are utilized to unify the node presentations from two view created by stochastic perturbation. Multi-view contrastive loss demonstrates outstanding outcomes facing interference and noise. 
Extensive experiments show that CSGDN achieves state-of-the-art performance, and outperforms baselines. Case study illustrate the superior performance of our model on crop datasets. CSGDN can predict positive/negative/irrelevant associations between gene and phenotype, and predictions are largely correct using TWAS results as a reference. As a result, CSGDN significantly improve the previously mentioned two issues and demonstrate essential biological significance.
In addition to TWAS results, the model can also accept other types of data inputs. For instance, researchers engaged in functional genomics can input gene-phenotype associations obtained from CRISPR, RNAi, or overexpression experiments into the CSGDN. This allows the model to provide new candidate genes for researchers in these fields.
However, the current model lacks interpretability, which limits its direct application in crop breeding. In the future, we look forward to applying advanced mechanisms to improve the interpretability to provide clearer insights into underlying gene-phenotype associations.

\section{ACKNOWLEDGEMENTS}
This study was supported by the Finance science and technology project of Xinjiang Uyghur Autonomous Region (2023A01). This study was also supported by the National Key Research and Development Program of China (2021YFF1000900) and the National Natural Science Foundation of China (W2411020, 32170645). We thank the high-performance computing platform at the National Key Laboratory of Crop Genetic Improvement in Huazhong Agricultural University.

\bibliographystyle{unsrt}
\bibliography{ref}




\end{document}